\documentclass[11pt]{article}

\usepackage{tikz}
\usetikzlibrary{positioning}
\usepackage{acl}
\usepackage[T1]{fontenc}

\title{Mirror, Mirror on the Wall: Prompt Echoing in Small Instruct Language Models}

\author{
 \textbf{Inez Okulska\textsuperscript{1}},
 \textbf{Bartosz Naskręcki\textsuperscript{1,2}},
 \textbf{Jan Piotrowski\textsuperscript{1}},
 \textbf{Tomasz Steifer\textsuperscript{1,3}},
\\
 \textsuperscript{1}Centre for Credible AI, Warsaw University of Technology,\\
 \textsuperscript{2}Adam Mickiewicz University Poznan,\\
 \textsuperscript{3}Institute of Fundamental Technological Research, Polish Academy of Sciences.
\\
 \small{
   \textbf{Correspondence:} \href{mailto:
anakin65@gmail.com}{anakin65@gmail.com}
 }
}

\definecolor{cQwen}{HTML}{2C7BB6}
\definecolor{cLlama}{HTML}{1A9850}
\definecolor{cGemma}{HTML}{D73027}
\definecolor{cOlmo1}{HTML}{F46D43}
\definecolor{cOlmo13}{HTML}{7B3294}
\definecolor{cSmol}{HTML}{008C8C}

\usepackage{booktabs}
\begin{document}
\maketitle
\begin{abstract}
Prompt echoing is a recognized failure mode of instruct language models, in which a model instead of generating a response, mirrors the provided prompt, even though it did not receive a specific instruction to do so. Is this phenomenon a sign of the model leaking the content of its training dataset, or is it rather caused by a misaligned behavior of the internal induction/copying mechanisms? We investigate prompt echoing small language models from different families (Gemma, Llama, Qwen, SmolLM and OLMo) and show that echoing prompts are likely to have partial overlap with the training dataset but the phenomenon is primarily driven by the model's induction heads.
\end{abstract}

Modern language models exhibit a wide range of failure modes, from mundane breakdowns in instruction following, formatting, and calibration to adversarially induced behaviors such as prompt injection and jailbreaking. The latter are now widely recognized as central security concerns. Beyond these adversarial settings, however, models can also fail in ways that are not caused by an explicit attack. One such failure is \textbf{prompt echoing}, where the model responds by copying the input prompt, or a long prefix/span of it, despite receiving no instruction to do so.

Prompt echoing has appeared in prior evaluations as a recognizable class of generation error. For example, \citet{seo2024evaluation} classify ``prompt echoing error'' as outputs that directly reflect the content of the prompt, and give examples where the model repeats prompt instructions instead of producing the intended concise response. More recently, evaluations of extended rule following also reported prompt echoes among natural failure modes~\citep{dai2026extended}. We use the term in this narrower sense and distinguish it from other recent use of ``echo'' in the literature, namely the intentional prompt-repetition techniques that improve answer quality~\citep{leviathan2025promptrepetitionimprovesnonreasoning,haoEchoesAnchorsProbabilistic2026,mekalaEchoPromptInstructingModel2024}. 

In this work, we study prompt echoing in small language models to understand its underlying mechanisms: rather than treating it only as a surface-level decoding artifact, we ask what internal model behaviors make echoing likely, and in particular, whether it is driven more by memorization or internal copying mechanisms.


\definecolor{EchoBlue}{HTML}{1D4ED8}
\definecolor{EchoBlueBg}{HTML}{EFF6FF}
\definecolor{NonEchoAmber}{HTML}{B45309}
\definecolor{NonEchoBg}{HTML}{FFFBEB}
\definecolor{BoxRule}{HTML}{CBD5E1}

\begin{figure}[t]
\centering
\begin{tikzpicture}[
  examplebox/.style={
    draw=BoxRule,
    rounded corners=1.2mm,
    line width=0.35pt,
    inner sep=3.6pt,
    text width=0.94\columnwidth,
    align=left
  },
  fullbox/.style={examplebox, fill=EchoBlueBg},
  nonbox/.style={examplebox, fill=NonEchoBg}
]

\node[fullbox] (recipe) {
  {\scriptsize\bfseries SmolLM2 1.7B Q4\_K\_M}
  \hfill {\scriptsize\bfseries\textcolor{EchoBlue}{full echo}}\par\vspace{0.4mm}
  {\fontsize{6.5}{7.1}\selectfont
  \begin{tabular}{@{}p{0.2\columnwidth}@{}p{0.79\columnwidth}@{}}
  \textbf{Prompt:} & To convert 2 cups of all-purpose flour to ounces, we multiply 2 cups by 8 ounces per cup. 2 * 8 = 16 ounces.\\[-0.25mm]
  \textbf{Answer:} & To convert 2 cups of all-purpose flour to ounces, we multiply 2 cups by 8 ounces per cup. 2 * 8 = 16 ounces.
  \end{tabular}}
};

\node[fullbox, below=1.4mm of recipe] (codeecho) {
  {\scriptsize\bfseries OLMo 0425 1B}
  \hfill {\scriptsize\bfseries\textcolor{EchoBlue}{full echo}}\par\vspace{0.4mm}
  {\fontsize{6.5}{7.1}\selectfont
  \begin{tabular}{@{}p{0.2\columnwidth}@{}p{0.79\columnwidth}@{}}
  \textbf{Prompt:} & \texttt{Thank you for your feedback! It's always appreciated. Don't hesitate to reach out if you need any more help. Enjoy your trip planning!}\\[-0.25mm]
  \textbf{Answer:} & \texttt{Thank you for your feedback! It's always appreciated. Don't hesitate to reach out if you need any more help. Enjoy your trip planning!}
  \end{tabular}}
};

\node[nonbox, below=1.4mm of codeecho] (nonecho) {
  {\scriptsize\bfseries SmolLM2 1.7B}
  \hfill {\scriptsize\bfseries\textcolor{NonEchoAmber}{non-echo}}\par\vspace{0.4mm}
  {\fontsize{6.5}{7.1}\selectfont
  \begin{tabular}{@{}p{0.2\columnwidth}@{}p{0.79\columnwidth}@{}}
  \textbf{Prompt:} & \texttt{\{ "fibonacci\_config": \{ "sequence\_limit": 10, "start\_number": 0, "generate\_numbers": true \} \}}\\[-0.25mm]
  \textbf{Answer:} & I'm sorry, but I couldn't understand your request. Could you please clarify what actions you need help with, such as generating a Fibonacci sequence, solving a specific math problem, or something else?
  \end{tabular}}
};

\end{tikzpicture}
\caption{
Two examples of prompt echoing and one of non-echo answer in which the model asks for clarification.
}
\label{fig:prompt-answer-examples-onecol}
\end{figure}

\paragraph{Our contribution}
We investigate prompt echoing in several small language models from different families: Qwen~\citep{yang2025qwen3,qwen2026qwen35_4b},
Llama~\citep{grattafiori2024llama3,meta2024llama32_3b},
Gemma~\citep{gemmateam2024gemma,google2025gemma3n_e4b},
OLMo~\citep{groeneveld-etal-2024-olmo},
and SmolLM~\citep{allal2025smollm2}. Our general working hypothesis is that prompt echoing is a combination of text memorization induced in pretraining and finetuning and an intrinsic copying mechanism existing in modern language models. We provide empirical evidence for the following observations:
\paragraph{1.}Prompt echoing happens across a diverse collection of language models, even under high temperature decoding (see Experiment 1).

\paragraph{2.} Echogenic prompts are likely to have partial overlap with texts from the training dataset. We confirm this by investigating public training datasets available for OLMo and SmolLM models (see Experiment 2).
\paragraph{3.}Once the echo is initialized, the continuation is carried out by the internal copying mechanism, akin to induction heads \cite{olsson2022induction}. We confirm that by forcing echo on a dataset of non-echogenic prompts across several models (see Experiment 3).
\paragraph{4.}Ablating attention heads with high copy scores significantly reduces echo across all models (see Experiment 4).




\paragraph{AI tools usage:} This research was facilitated by responsible usage of LLM tools. In particular, ChatGPT and codex were used preparation of parts of the code as well as manuscript drafting, including LaTeX figure preparation, grammar editing and proof reading. The results and claims were independently verified by the authors. The experiments were conceived and designed by the authors. We take full responsibility for the context of the paper.

\paragraph{Related work}

\citet{olsson2022induction} identified induction heads as attention heads implementing a prefix-matching and copying algorithm. These authors hypothesized that induction heads are a necessary component for in-context learning. \citet{mcdougall-etal-2024-copy} characterize a class of \emph{copy-suppression} heads whose role is to attenuate, rather than amplify, the in-context copying signal.




\paragraph{Memorization and repetition}

\citet{carliniextracting} showed that language models reproduce verbatim spans of their pretraining data, and \citet{carlini2023quantifyingmemorizationneurallanguage} establish log-linear relationships between memorization rate and model capacity, the number of duplicates in the training data, and the length of the prompting context. 


\section{Experimental setup}

\paragraph{Reproducibility} All the experiments were run on a cluster utilizing NVIDIA GH200 superchips. The code was written in Python and used PyTorch/Hugging Face Transformers for full-precision model inference, and llama.cpp for official GGUF quantized models. The datasets, download scripts, and evaluation code are publicly available at \href{https://github.com/CredibleAI/Echo}{https://github.com/CredibleAI/Echo}.

\subsection{Data preparation.} 
For our experiments we have constructed three sets of prompts: two datasets of echogenic prompts and one dataset of non-echogenic prompts.

\paragraph{Dataset A} For each source model, we first asked the model itself to generate standalone text under a diverse set of lightweight tasks, e.g. explanatory paragraphs, code snippets, mathematical explanations, documentation-style examples, JSON-like configurations, tutorials, and problem statements. The generated text was then treated as a candidate prompt. We filtered out candidates that were too short (less than 24 tokens) or too long (more than 192 tokens).

Each candidate was then screened by prompting the same model with that candidate text and sampling 10 continuations using non-greedy decoding (temperature 0.9). A candidate was retained if at least one sampled continuation contained a contiguous span of at least 40\% of the prompt tokens. We repeated this procedure until we obtained 64 accepted prompts for each of six source models. This yielded a 384-prompt dataset with known source model labels.

\paragraph{Dataset B} was constructed using recursive loop of each model response becoming its next prompt. For each of the six models, we initialized 64 independent chains with short synthetic user requests drawn from a fixed set of topic/task templates. In the base step, the generated text was fed to the model. At each following step, the model's last answer was given back to the model as a fresh prompt. This was continued for 10 steps. The model had no access to the history of the conversation, each prompt was starting a new interaction of prompt-answer duet. At each round we sampled a maximum of 180 new tokens. The tenth answer from each chain was saved as the final prompt, yielding 384 prompts total. Because in most cases the echo emerged not in the first round of prompt-response iterations, but in the next ones, which, given the recursive framework of generating the prompts from model's responses, caused the rather unusual conversation type, where the model is not asked, prompted or told to execute any action but confronted with a statement or re-action typical for a "helpful assistant", a role usually reserved for the model itself. 

\paragraph{Dataset C} was the reference dataset of non-echoing prompts. Using the same self-generation pipeline as for Dataset A, we iteratively generated text, testing each one for echogenicity. The screening condition was the negation of the previous one: a prompt was accepted only if none of the screened completions produced a prefix echo. The dataset consists of 128 prompts.  

\begin{table*}[!t]
\centering
\scriptsize
\setlength{\tabcolsep}{4pt}
\begin{tabular}{llrrrrrr}
\toprule
\textbf{Model} & \textbf{Temp.} & \multicolumn{3}{c}{\textbf{Dataset A}} & \multicolumn{3}{c}{\textbf{Dataset B}} \\
\cmidrule(lr){3-5}\cmidrule(lr){6-8}
 & & \textbf{Full} & \textbf{Partial} & \textbf{Any echo} & \textbf{Full} & \textbf{Partial} & \textbf{Any echo} \\
\midrule
Qwen3.5 4B & 1.0 & 0.01\% & 0.07\% & 0.08\% & 0.00\% & 0.01\% & 0.01\% \\
 & 0.3 & 0.00\% & 0.00\% & 0.00\% & 0.26\% & 0.00\% & 0.26\% \\
 & 0.0 & 0.00\% & 0.00\% & 0.00\% & 0.52\% & 0.00\% & 0.52\% \\
\midrule
Llama 3.2 3B & 1.0 & 0.00\% & 0.05\% & 0.05\% & 0.01\% & 1.12\% & 1.13\% \\
 & 0.3 & 0.52\% & 0.00\% & 0.52\% & 2.08\% & 1.56\% & 3.65\% \\
 & 0.0 & 0.26\% & 0.00\% & 0.26\% & 2.60\% & 1.04\% & 3.65\% \\
\midrule
Gemma 3n E4B & 1.0 & 0.00\% & 1.03\% & 1.03\% & 0.00\% & 8.95\% & 8.95\% \\
 & 0.3 & 1.30\% & 0.78\% & 2.08\% & 8.59\% & 0.78\% & 9.38\% \\
 & 0.0 & 1.56\% & 0.78\% & 2.34\% & 8.33\% & 0.78\% & 9.11\% \\
\midrule
OLMo-2 0425 1B & 1.0 & 0.14\% & 0.79\% & 0.94\% & 0.09\% & 0.18\% & 0.27\% \\
 & 0.3 & 0.26\% & 0.00\% & 0.26\% & 1.30\% & 0.00\% & 1.30\% \\
 & 0.0 & 0.26\% & 0.00\% & 0.26\% & 1.56\% & 0.00\% & 1.56\% \\
\midrule
OLMo-2 1124 13B & 1.0 & 0.00\% & 0.38\% & 0.38\% & 0.13\% & 0.16\% & 0.29\% \\
 & 0.3 & 0.26\% & 0.26\% & 0.52\% & 0.52\% & 0.00\% & 0.52\% \\
 & 0.0 & 0.52\% & 0.52\% & 1.04\% & 1.30\% & 0.26\% & 1.56\% \\
\midrule
SmolLM2 1.7B & 1.0 & 0.59\% & 1.04\% & 1.63\% & 0.43\% & 1.84\% & 2.27\% \\
 & 0.3 & 3.39\% & 1.30\% & 4.69\% & 8.85\% & 1.30\% & 10.16\% \\
 & 0.0 & 5.21\% & 0.52\% & 5.73\% & 11.46\% & 1.04\% & 12.50\% \\
\midrule
OLMo-2 0425 1B Q4\_K\_M & 1.0 & 0.05\% & 0.44\% & 0.49\% & 0.08\% & 0.20\% & 0.27\% \\
 & 0.3 & 0.52\% & 0.52\% & 1.04\% & 0.26\% & 0.00\% & 0.26\% \\
 & 0.0 & 0.78\% & 0.00\% & 0.78\% & 1.04\% & 0.00\% & 1.04\% \\
\midrule
OLMo-2 1124 13B Q4\_K\_M & 1.0 & 0.22\% & 0.87\% & 1.09\% & 0.38\% & 0.51\% & 0.89\% \\
 & 0.3 & 1.04\% & 0.78\% & 1.82\% & 0.78\% & 0.26\% & 1.04\% \\
 & 0.0 & 1.04\% & 0.26\% & 1.30\% & 0.52\% & 0.26\% & 0.78\% \\
\midrule
SmolLM2 1.7B Q4\_K\_M & 1.0 & 1.80\% & 3.07\% & 4.87\% & 0.91\% & 4.60\% & 5.51\% \\
 & 0.3 & 7.81\% & 2.34\% & 10.16\% & 10.94\% & 1.56\% & 12.50\% \\
 & 0.0 & 8.07\% & 2.08\% & 10.16\% & 13.54\% & 1.82\% & 15.36\% \\
\bottomrule
\end{tabular}
\caption{Experiment 1 prefix-echo rates across temperatures. For T=1.0, each model-prompt pair was sampled 20 times; for T=0.3 and T=0.0, only once. Full echo means the continuation begins with the complete prompt; partial echo means it begins with at least 40\% of the prompt tokens but diverges before copying the full prompt.}
\label{tab:experiment1-echo-rates}
\end{table*}


\begin{table*}[h!]
\centering
\scriptsize
\setlength{\tabcolsep}{3.0pt}
\begin{tabular}{ll*{9}{r}}
\toprule
\textbf{Model} & \textbf{Variant} &
\multicolumn{3}{c}{\textbf{$k=2$ forced tokens}} &
\multicolumn{3}{c}{\textbf{$k=3$ forced tokens}} &
\multicolumn{3}{c}{\textbf{$k=5$ forced tokens}} \\
\cmidrule(lr){3-5}
\cmidrule(lr){6-8}
\cmidrule(lr){9-11}
& & \textbf{Full} & \textbf{Partial} & \textbf{Rate}
& \textbf{Full} & \textbf{Partial} & \textbf{Rate}
& \textbf{Full} & \textbf{Partial} & \textbf{Rate} \\
\midrule
Gemma 3n E4B IT & full
& 71 & 341 & 16.1\%
& 229 & 565 & 31.0\%
& 284 & 738 & 39.9\% \\
Llama 3.2 3B Instruct & full
& 58 & 80 & 5.4\%
& 93 & 178 & 10.6\%
& 111 & 407 & 20.2\% \\
OLMo-2 0425 1B Instruct & full
& 3 & 40 & 1.7\%
& 4 & 112 & 4.5\%
& 8 & 189 & 7.7\% \\
OLMo-2 0425 1B Instruct & Q4\_K\_M
& 16 & 89 & 4.1\%
& 37 & 407 & 17.3\%
& 40 & 570 & 23.8\% \\
OLMo-2 1124 13B Instruct & full
& 0 & 45 & 1.8\%
& 3 & 169 & 6.7\%
& 3 & 200 & 7.9\% \\
OLMo-2 1124 13B Instruct & Q4\_K\_M
& 12 & 160 & 6.7\%
& 46 & 382 & 16.7\%
& 59 & 499 & 21.8\% \\
Qwen3.5 4B & full
& 115 & 211 & 12.7\%
& 195 & 426 & 24.3\%
& 322 & 697 & 39.8\% \\
SmolLM2 1.7B Instruct & full
& 65 & 234 & 11.7\%
& 128 & 391 & 20.3\%
& 133 & 536 & 26.1\% \\
SmolLM2 1.7B Instruct & Q4\_K\_M
& 159 & 411 & 22.3\%
& 253 & 529 & 30.5\%
& 271 & 636 & 35.4\% \\
\bottomrule
\end{tabular}
\caption{
Effect of inducing echo by forcing the assistant's answer to begin with prefix of the prompt.
}
\label{tab:experiment3-seeded-echo}
\end{table*}
 

\begin{table*}[h!]
\centering
\scriptsize
\setlength{\tabcolsep}{2.4pt}
\begin{tabular}{l*{12}{r}}
\toprule
\textbf{Model} &
\multicolumn{3}{c}{\textbf{No ablation}} &
\multicolumn{3}{c}{\textbf{Top-8 copy heads}} &
\multicolumn{3}{c}{\textbf{Top-16 copy heads}} &
\multicolumn{3}{c}{\textbf{Random-16 heads}} \\
\cmidrule(lr){2-4}
\cmidrule(lr){5-7}
\cmidrule(lr){8-10}
\cmidrule(lr){11-13}
& \textbf{Full} & \textbf{Partial} & \textbf{Rate}
& \textbf{Full} & \textbf{Partial} & \textbf{Rate}
& \textbf{Full} & \textbf{Partial} & \textbf{Rate}
& \textbf{Full} & \textbf{Partial} & \textbf{Rate} \\
\midrule
Qwen3.5 4B Instruct
& 2 & 8 & 0.07\%
& 0 & 1 & 0.01\%
& \textbf{0} & \textbf{0} & 0.00\%
& 1 & 4 & 0.03\% \\
Llama 3.2 3B Instruct
& 2 & 94 & 0.62\%
& 1 & 6 & 0.05\%
& 0 & 2 & 0.01\%
& 1 & 42 & 0.28\% \\
Gemma 3n E4B IT
& 0 & 764 & 4.97\%
& 0 & 1 & 0.01\%
& \textbf{0} & \textbf{0} & 0.00\%
& 0 & 715 & 4.65\% \\
OLMo-2 0425 1B Instruct
& 16 & 93 & 0.71\%
& 1 & 0 & 0.01\%
& \textbf{0} & \textbf{0} & 0.00\%
& 4 & 82 & 0.56\% \\
OLMo-2 1124 13B Instruct
& 8 & 45 & 0.35\%
& 0 & 6 & 0.04\%
& 0 & 2 & 0.01\%
& 10 & 44 & 0.35\% \\
SmolLM2 1.7B Instruct
& 79 & 223 & 1.97\%
& 0 & 11 & 0.07\%
& 0 & 6 & 0.04\%
& 64 & 162 & 1.47\% \\
\bottomrule
\end{tabular}
\caption{
Effect of ablating attention heads with high prefix-match score
}
\label{tab:experiment5-head-ablation}
\end{table*}

\paragraph{Experiment 1: Echo rates}
In the first experiment, we have evaluated datasets A and B on the following models: Qwen 3.5 4B, Llama 3.2 3B, Gemma 3n E4B IT, OLMo-2 0425 1B, OLMo-2 1124 13B, and SmolLM2 1.7B, as well as on an additional batch of quantized models: OLMo-2 0425 1B Q4\_K\_M, OLMo-2 1124 13B Q4\_K\_M and SmolLM2 1.7B Q4\_K\_M. All models were Instruct variants. Each target model was prompted with every prompt in the dataset, independent of which model originally generated the prompt. For the T=1.0 condition, each model-prompt pair was sampled 20 times using non-greedy decoding with top\_p=1.0, top\_k=0\footnote{Top-p sampling restricts each decoding step to the smallest set of next-token candidates whose cumulative probability mass reaches $p$; top\_p=1.0 therefore leaves the full probability distribution available. Top-k sampling restricts generation to the $k$ highest-probability tokens; under the decoding convention used here, top\_k=0 disables this rank-based filter. Thus, the T=1.0 condition uses temperature-scaled sampling without additional top-p or top-k truncation.}, and a maximum of 160 new tokens. To test whether echoing depends on high-entropy sampling, we also evaluated temperatures 0.0 and 0.3 for the target models shown in Table~\ref{tab:experiment1-echo-rates}, using the same prompt set, one completion per model-prompt pair, and the same prefix-based echo labels.

Echoes were labeled using a strict prefix-only criterion. Therefore, a completion was counted as a full echo only then, if the generated continuation began with the complete prompt token sequence under the evaluated model tokenizer. A completion was counted as a partial echo if it began with at least 40\% of the prompt tokens but diverged before copying the full prompt. This definition intentionally excludes ordinary answers that quote or reuse the prompt later in the response.

\paragraph{Experiment 2: Training-Data Overlap Scan}
Experiment 2 asks whether the echo-prone prompts resemble text present in public training corpora associated with a subset of the studied model, namely, instruction dataset SmolTalk and Tulu-3 SFT mixture for OLMo-2. For each of the prompt scoring at least a partial echo in Experiment 1 on SmolLM or OLMo models, we look for partial overlap in their respective public training datasets. A match was recorded if the first 5 words of the prompt occur contiguously anywhere inside a dataset document. Note that this is a weak criterion that should not be interpreted as a sign of strong training data leakage. 


\paragraph{Experiment 3: Forcing echo}
Experiment 3 tests whether echoing can be induced even for prompts that do not echo spontaneously. For each prompt in the dataset C we simulated a scenario in which the model has already begun copying the prompt. This was achieved by forcing the assistant answer to start with the first $k$ prompt tokens. We evaluated $k \in \{2,3,5\}$, allowing us to measure how much initial copying is needed before the model's own continuation mechanism takes over. After the forced seed tokens, generation proceeded normally with sampled non-greedy decoding. 


\paragraph{Experiment 4: Ablating copying heads}

In the last experiment, we investigated the effect of the ablation of induction/copying heads. We repeat the Experiment 1 evaluation on both echogenic datasets, but intervene on selected attention heads during generation. For each model, heads are first ranked using the prefix-matching score of \cite{olsson2022induction}, computed on repeated random inputs. 

We then run the prompt-echo evaluation under four conditions: no intervention, ablation of the top 8 copying heads, ablation of the top 16 copying  heads, and ablation of 16 randomly selected heads as a control. Ablation is implemented by zeroing the selected heads' attention-output slices before the output projection during generation. Echoes are scored with the same strict prefix-only labels as in Experiment 1. The experiment was run for the full-precision models collection.

\begin{figure}[!t]
\centering
\scriptsize
\begin{tikzpicture}[
  heat/.style={minimum width=0.68cm,minimum height=0.35cm,inner sep=0pt,draw=white,line width=0.45pt,font=\scriptsize\bfseries},
  head/.style={font=\tiny,anchor=south,align=center},
  rowlabel/.style={font=\scriptsize,anchor=east},
  totalcell/.style={minimum width=0.62cm,minimum height=0.35cm,inner sep=0pt,draw=gray!45,fill=gray!10,font=\scriptsize\bfseries}
]

\node[font=\scriptsize,anchor=west] at (-0.96,2.08) {SmolLM2-target echoes by prompt source};
\node[head] at (0.00,1.48) {Qw};
\node[head] at (0.78,1.48) {Ll};
\node[head] at (1.56,1.48) {Ge};
\node[head] at (2.34,1.48) {O1};
\node[head] at (3.12,1.48) {O13};
\node[head] at (3.90,1.48) {Sm};
\node[head] at (4.78,1.48) {All};

\node[rowlabel] at (-0.48,1.10) {A, 0.0};
\node[heat,fill=blue!22!white,text=black] at (0.00,1.10) {6};
\node[heat,fill=blue!14!white,text=black] at (0.78,1.10) {4};
\node[heat,fill=blue!43!white,text=black] at (1.56,1.10) {12};
\node[heat,fill=blue!22!white,text=black] at (2.34,1.10) {6};
\node[heat,fill=blue!32!white,text=black] at (3.12,1.10) {9};
\node[heat,fill=blue!43!white,text=black] at (3.90,1.10) {12};
\node[totalcell] at (4.78,1.10) {49};

\node[rowlabel] at (-0.48,0.70) {A, 0.3};
\node[heat,fill=blue!14!white,text=black] at (0.00,0.70) {4};
\node[heat,fill=blue!11!white,text=black] at (0.78,0.70) {3};
\node[heat,fill=blue!40!white,text=black] at (1.56,0.70) {11};
\node[heat,fill=blue!43!white,text=black] at (2.34,0.70) {12};
\node[heat,fill=blue!32!white,text=black] at (3.12,0.70) {9};
\node[heat,fill=blue!32!white,text=black] at (3.90,0.70) {9};
\node[totalcell] at (4.78,0.70) {48};

\node[rowlabel] at (-0.48,0.30) {B, 0.0};
\node[heat,fill=blue!14!white,text=black] at (0.00,0.30) {4};
\node[heat,fill=blue!47!white,text=black] at (0.78,0.30) {13};
\node[heat,fill=blue!76!white,text=white] at (1.56,0.30) {21};
\node[heat,fill=blue!14!white,text=black] at (2.34,0.30) {4};
\node[heat,fill=blue!18!white,text=black] at (3.12,0.30) {5};
\node[heat,fill=blue!83!white,text=white] at (3.90,0.30) {23};
\node[totalcell] at (4.78,0.30) {70};

\node[rowlabel] at (-0.48,-0.10) {B, 0.3};
\node[heat,fill=blue!11!white,text=black] at (0.00,-0.10) {3};
\node[heat,fill=blue!36!white,text=black] at (0.78,-0.10) {10};
\node[heat,fill=blue!61!white,text=white] at (1.56,-0.10) {17};
\node[heat,fill=blue!14!white,text=black] at (2.34,-0.10) {4};
\node[heat,fill=blue!18!white,text=black] at (3.12,-0.10) {5};
\node[heat,fill=blue!86!white,text=white] at (3.90,-0.10) {24};
\node[totalcell] at (4.78,-0.10) {63};

\node[font=\tiny,anchor=east] at (1.24,-0.64) {0};
\foreach \x/\shade in {1.38/0,1.68/23,1.98/45,2.28/68,2.58/90}
  \node[minimum width=0.25cm,minimum height=0.13cm,inner sep=0pt,draw=gray!35,fill=blue!\shade!white] at (\x,-0.64) {};
\node[font=\tiny,anchor=west] at (2.74,-0.64) {25};

\end{tikzpicture}
\caption{Prompt-source heatmap for low-temperature echoes into the SmolLM2 target family. Cells count unique prompts, out of 384 prompts in the corresponding dataset, that produced a full or partial prefix echo in at least one SmolLM2 target model (fp32 or Q4\_K\_M). Source columns: Qw=Qwen, Ll=Llama, Ge=Gemma, O1=OLMo-2 0425 1B, O13=OLMo-2 1124 13B, Sm=SmolLM2.}
\label{fig:smollm2-transfer-source-composition}
\end{figure}

\section{Results and interpretation}

The results of \textbf{Experiment 1} are summarized in Table~\ref{tab:experiment1-echo-rates}. Prompt echoing occurs across all six model families, even at temperature~1.0. Quantization amplifies the effect for some models (SmolLM2 jumps from 1.95\% to 5.19\%) but not universally (OLMo-2 1B drops slightly from 0.61\% to 0.38\%). The lower-temperature conditions show that echoing does not vanish under greedy or near-greedy decoding. On the contrary, in most tested model-dataset combinations, echo rates are higher at T=0.0 or T=0.3 than at T=1.0. This pattern is most visible for Dataset B and for the SmolLM2 family, especially the Q4\_K\_M variant, where the any-echo rate reaches 15.36\% at T=0.0. We have also examined the source-model composition for the highest echoing model. Figure~\ref{fig:smollm2-transfer-source-composition} shows that SmolLM2-target echoes are not restricted to prompts originally generated by SmolLM2; in every low-temperature condition, most come from other source models.

In the \textbf{Experiment 2} we observed that 32.63\% of prompts generating echo in OLMo matched some text in the training dataset as compared to 9.38\% for non-echogenic prompts. The difference was weaker for SmolLM model---26.33\% for echoing prompts vs 17.97\% for non-echogenic prompts. This suggests memorization does not explains the echo well.

The results of \textbf{Experiment~3} are summarized in Table~\ref{tab:experiment3-seeded-echo}. The experiment confirms that forcing even 2~initial prompt tokens into the assistant's answer results in a significantly increased rate of partial and full echoes. For Gemma~3n~E4B~IT, 5~forced tokens increase the echo rate to almost 40\%, even under high-temperature decoding. This supports the hypothesis that once the echo kicks in, it is continued by the internal copying mechanism — even for prompts that were selected not to echo
spontaneously.
 
This is further supported by \textbf{Experiment~4}, in which we ablated the internal copying mechanism in the form of the model's induction heads. Ablation of candidate induction heads causes a larger drop in echo rates than ablating randomly chosen attention heads across all models. We also observe a larger decrease in echo rate when more candidate induction heads are ablated (16 rather than 8). This is consistent with previous work which reported that the copying mechanism is not uniquely localized in a single head but rather spread across several heads~\citep{olsson2022induction,bansal-etal-2023-rethinking,crosbie-shutova-2025-induction}. The results of Experiment~4 are summarized in Table~\ref{tab:experiment5-head-ablation}.

\paragraph{Conclusions} Prompt echo appears to arise from several interacting mechanisms: it is influenced by memorization but not fully explained by it, and once initialized, it is primarily driven by internal copying mechanisms. Its persistence at temperatures 0.0 and 0.3 further suggests that echoing is not only a rare tail-sampling outcome, but can become a stable decoding trajectory for some prompt-model pairs. Future work may isolate the echo-initiation step more directly, quantization-dependent shifts or activation patching and circuit discovery, to move beyond head-level ablations toward a mechanistic account.

\section*{Limitations}

Our study focuses on a narrow, prefix-based form of prompt echoing in small instruct-tuned language models, and the reported echo rates should be interpreted as measurements on diagnostic prompt distributions rather than estimates of deployment prevalence.
Our strict token-prefix definition makes echoing easy to score and analyze, but excludes related behaviors such as copying later spans of the prompt, paraphrasing the prompt, quoting it inside an otherwise normal answer, or producing more diffuse repetition. 

Our training-data overlap analysis is limited to only two models: public training or instruction-tuning data are available only for some model families. Short prefix matches do not by themselves establish verbatim memorization or training-data leakage. Similarly, our induction-head ablations test whether heads with high prefix-matching scores are causally involved in echoing, but head ablation is a coarse intervention and does not fully identify the underlying circuit. Finally, our experiments cover only a limited set of English, one-shot prompts and small instruct models; the conclusions may not directly transfer to larger models, multilingual settings, long-context use, or multi-turn interactions.

 \section*{Acknowledgements}
Work on this project is financially supported by the Foundation for Polish Science (FNP) grant
‘Centre for Credible AI’ No. FENG.02.01-IP.05-0058/24.

\bibliography{custom}

\appendix
\newpage


\end{document}